\documentclass[letterpaper, 10 pt, conference]{ieeeconf}  

\usepackage{gensymb}
\usepackage{bm}
\usepackage{caption}
\usepackage{cite}
\usepackage{framed}
\usepackage{sidecap}
\usepackage{graphicx}
\graphicspath{ {./figs/} }
\usepackage{xcolor}
\usepackage{hyperref}
\hypersetup{
    colorlinks=true,
    linkcolor=magenta,
    filecolor=forestgreen,      
    urlcolor=cyan,
    pdftitle={Overleaf Example},
    pdfpagemode=FullScreen,
    }

\usepackage{marginnote}
\usepackage{soul}

\IEEEoverridecommandlockouts                              

\title{\LARGE \bf Can Robotic Experimenters help improve HRI Experiments?\\
An Experimental Study}

\author{Dan R. Suissa$^{1*}$, Shikhar Kumar $^{2*}$ and Yael Edan$^{2}$  
\thanks{$^{1}$ Dan is with the Department of Computer Science, Ben-Gurion University of the Negev, Be'er Sheva, Israel {\tt\small danrouve@bgu.ac.il, dan.rouven.suissa@gmail.com}}%
\thanks{$^{2}$Shikhar and Yael are with the Department of Industrial Engineering and Management, Ben-Gurion University of the Negev, Be'er Sheva, Israel 
{\tt\small shikhar and yael@bgu.ac.il}}%
\thanks{$^*$ Equal Contribution}
}

\begin{document}
\maketitle
\thispagestyle{empty}
\pagestyle{empty}

\begin{abstract}
To evaluate the design and skills of a robot or an algorithm for robotics, human-robot interaction user studies need to be performed.  
Classically, these studies are conducted by human experimenters, requiring considerable effort, and introducing variability and potential human error. 
In this paper, we investigate the use of robots in support of HRI experiments.
Robots can perform repeated tasks accurately, thereby reducing human effort and improving  validity through reduction of error and variability between participants.
To assess the potential for robot-led HRI experiments, we ran an HRI experiment with two participant groups – one led by a human experimenter and another led mostly by a robot experimenter.
We show that the replacement of several repetitive experiment tasks through robots is not only possible but beneficial:
Trials performed by the robot experimenter had fewer errors and were more fluent.
There was no statistically significant difference in participants' perception w.r.t. cognitive load, comfortability, enjoyment, safety, trust and understandability between both groups. 
To the best of our knowledge, this is the first comparison between robot-led and human-led HRI experiments. 
It suggests that using robot experimenters can be beneficial and should be considered.\\

{\em Keywords---}
Human-Robot Interaction; Autonomous Experiments; Human-Centered Robotics; Acceptability, Trust, Fluency;
\end{abstract}


\section{Introduction}
The domain of human-robot interaction (HRI) involves designing, evaluating and understanding the usage of robots, through experiments \cite{HCI-005} following rigorous methodology to obtain knowledge \cite{hempel1966philosophy}.
To evaluate the design and skills of a particular robot or function of an algorithm, often, a user study needs to be performed. 
A strict procedure has to be followed from the time participants start the experiment to the end of it \cite{hoffman2020primer}.
Experimenters in HRI studies must precisely follow the \textit{"experimental scripts"} to remove variability from their influence on the results as much as possible.

These studies are classically conducted with a human as the experimenter \cite{hoffman2020primer} and examples are manifold, e.g., \cite{kumar2022politeness, singh2021verbal, dennler2022design, kshirsagar2020robot}.
Conducting a user experiment requires repetitive actions on the experimenter part.
For example: the greeting of participants, filling of forms and questionnaires, instructions regarding purpose and background of the experiment and its procedure, starting the experiment (robots and other technology), observing it, taking notes, ensuring its safety, and so on \cite{hoffman2020primer}.
This is obviously a lot of effort and tedious work.
More importantly, we humans are bound to make mistakes in these repetitive tasks and they are under influence of human-human interactions.
Also, if experiments are performed on consecutive days or are to be repeated at other times and locations, the experiment conditions will very likely vary. 
This is unfortunate, as it is important to ensure identical experimental conditions for all participants.

In an effort to improve consistency and validity, and to reduce human effort, we investigated adding robotic experimenters to support us while conducting an HRI study.

An autonomous robot experimenter could have several benefits: 
\begin{itemize}
    \item potential to make less mistakes
    \item reduces variability between trials
    \item reduce human-human influence
    \item allows for precise replicability\footnote{at any time and location with suitable robots}
    \item allows for sharing of experiments\footnote{directly via code}
    \item reducing effort for human experimenters\\
\end{itemize}

To this end, we compare the influence of performing the repetitive tasks of the experiment with a robotic experimenter.
We repeat the same HRI experiment, once led by a human experimenter and once led mostly by a scripted autonomous robot experimenter.
To the best of our knowledge, this is the first time that a direct comparison of human and robot experimenters for HRI studies is conducted.

\section{AUTOMATING HUMAN-ROBOT INTERACTION EXPERIMENTS}
The general idea to automate experiments through robotics is not new, however in HRI the idea is under-explored with respect to the automation of experiment tasks and the experiment procedure.
We only found one study where two robots were recruiting bystanders as audience to watch and rate the comedic value of their comedy show \cite{swaminathan2021robots}.
The robots conducted their own human experiments in that they queried the audience after displaying experimentally balanced episodes of relational humor.
This study partially shows the viability of robots conducting their own experiments by recruiting participants off the street.
However, it does not treat a structured HRI experiment, there is no experimenter and thus it does not test for and discuss the automation of HRI experiments through robotics.

This leads us to our proposal, for which we first discuss the typical procedure of HRI experiments \cite{hoffman2020primer}:
After initial greetings, participants are usually filling consent forms. 
After their consent, often participants need to fill further forms and a preliminary questionnaire, which inform on their background and biases.
If needed, participants are equipped with physiological measuring devices before they move towards the experiment area, and usually they are recorded with video and audio during the experiment.
Then after completion of (a part of) the experiment the participants fill further questionnaires, which are now related to the experiment.
This might be repeated several times.
The experimenter finally conducts an interview and debriefs the participants. 
Note, that some experiments require the same participants to come multiple times and experience an experiment consecutively or with different setting.

We propose, that most of the above steps that are repeated during each experiment, can be automated using robots.
This frees up us humans conducting the experiment to focus on observation tasks rather than repetitive explanations and procedure.
In our experiments we automated all these tasks, leaving us with only: the greeting, consent and a final debrief - all outside the lab (outside the experiment space).

An automated HRI experiment includes the actor robot(s) of the HRI experiment itself and at least one experimenter robot which supports the human experimenter(s).
Note, that sometimes the actor robot can also fulfill the role of the experimenter robot (like in \cite{swaminathan2021robots}).
Also note, that the influence of having a robot run its own experiment versus adding robots to run the experiment (our approach) is beyond the scope of what we review in this study.

\section{METHODS}
The comparative experiments took place in the Autonomous Multi Robot Lab in the Industrial Engineering Department.
A scene from the experiment and the experimentation area layout can be seen in Fig. \ref{fig:1}.

\subsection{Experimental Design}
\begin{figure}[h!]
\centering
\vspace{0.15cm}
\includegraphics[width=0.35\textwidth]{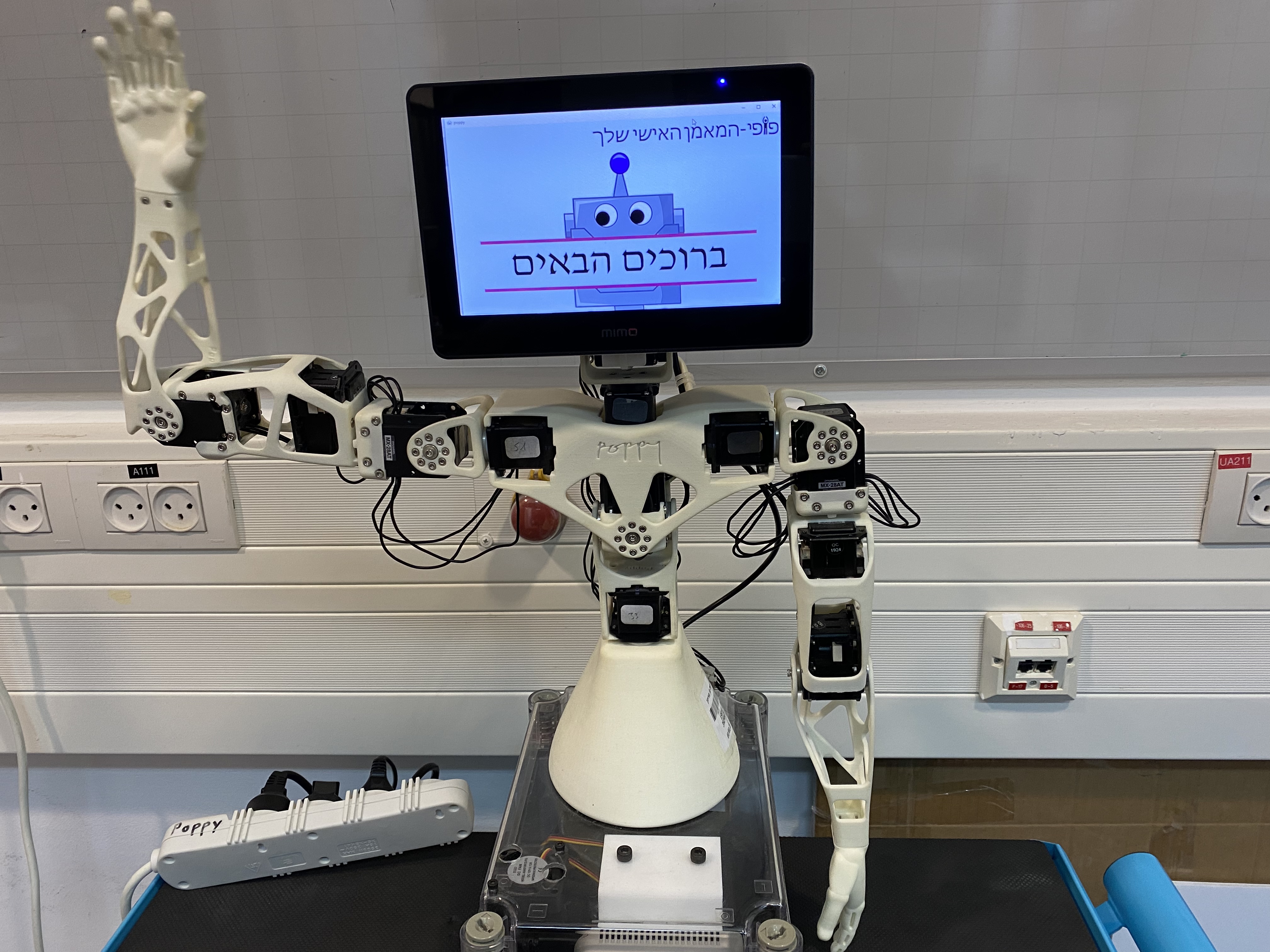}
\caption{Picture of the {\em Gymmy} physical trainer as used in the training experiment.}
\label{fig:2}
\end{figure}

We conducted a between-groups user study, comparing a human experimenter (HE) with a robot experimenter (RE).
This comparative study was build around a rather simple HRI experiment, identical for both groups, it included a short physical training session with the {\em Gymmy} robot~\cite{krakovski2021gymmy, avioz2021robotic} (see Fig. \ref{fig:2}).
Participants in both groups (HE and RE) were not informed about the existence of groups or the comparative experiment.
Both groups were greeted by us, signed consent and then entered a lab in which they would either meet HE or RE. 
After the experiment within the lab, they would exit, fill a final questionnaire and get a short debrief (not to inform other students on the nature of their experiment). 
This means that for all students, the experiment took place in a separate space to the greeting area and for RE, during the stay in the lab, this would mean not observing any other humans. 
We opted for a non-humanoid, mobile robot with speakers and a screen as RE (Wyca {\em Keylo} tele-presence robot).
A research assistant who was compensated by the university, was recruited as HE.
The experiments were conducted on three consecutive days (from 10am to 4pm with a 1h lunch break).
A single experiment would last on average less than 20 minutes.
On the first day the HE would start and on the second the RE -- they would switch after lunch.
On day three we performed four more HE experiments before lunch.

During all experiments, for observation and safety reasons, a further human observer was present, unnoticed by the participants (within the lab).

\begin{figure}[h!]
\centering
\includegraphics[width=0.45\textwidth]{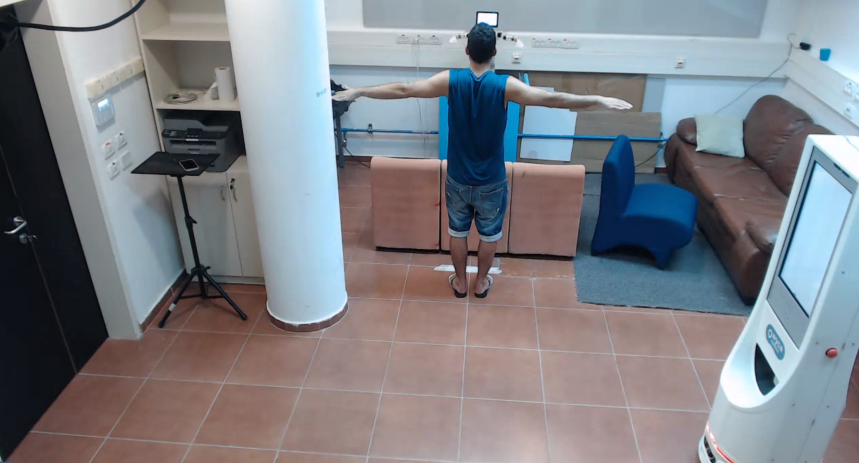}
\caption{Picture of the autonomous HRI environment. With the robot experimenter {\em Keylo} on the right hand side and a participant training with {\em Gymmy}, the robotic physical trainer (in the training experiment).}
\label{fig:1}
\end{figure}

During the stay in the experiment area (lab), the procedure was the same for both groups (HE, RE).
First, participants would enter the lab where the experimenter welcomed them with a pre-defined text.
Then, the experimenter guided towards a computer where participants filled the pre-experiment questionnaire.
All questionnaires, were provided using Google forms.
See supplementary material%
\footnote{\url{https://github.com/D-R-S/ahri_supplement.git}}, for questionnaires.
After completion of the first questionnaire, the experimenter instructed the participants about the HRI experiment and start the {\em Gymmy} robot.
Then the training would take place (see Fig. \ref{fig:1}, participant training with Gymmy).
After finishing the training experiment (see next subsection), the experimenter would guide the participants back to the questionnaire computer.
Participants would fill a first post experiment questionnaire (denoted "internal").
This questionnaire was rating the interaction / training with the {\em Gymmy} robot.
Then, the experimenter would thank the participants and send them out of the lab -- concluding the HRI experiment.
Finally, participants would fill a last post experiment questionnaire once outside (denoted "external").
The participants were instructed that this questionnaire was about the whole experience inside the lab.
Thus they rated the interaction with the experimenter (HE or RE) and the procedure itself concluding the comparative experiment.

\subsection{Gymmy Training Experiment}
We simplified and repeated an experiment conducted with {\em Gymmy}, a physical training robot, which was originally designed for older adults \cite{krakovski2021gymmy}.
In the original experiment, participants would absolve a physical training with feedback and a cognitive game.
We removed the cognitive game from the experiment as well as the feedback the robot would give the participants.
This left us with a five minute experiment of physical training of the upper body.
It included five exercises with one set of eight repetitions each.
We chose this experiment of reduced complexity to ensure that each participant would perceive the training experiment in a similar way -- simple and quick.
Thus, the completion of the internal experiment would have a similar influence on the overall study for all participants.
Besides that, the chosen experiment had a high probability of success\footnote{In our experiments the training HRI experiment showed a success rate of 1.0}.
We collected data via questionnaires, videos and in addition, the {\em Gymmy} robot acquired automatically video and motion capture data.
This data can inform on the use of the Gymmy robot with younger adults (instead of the elderly for which it was originally purposed).
It is important to note, that both the physical training as well as the cognitive game we omitted, are perceived as trivially easy by most (healthy) younger adults -- which we were aware of beforehand and based on which we opted for this experiment.

\subsection{Robot Experimenter}
\begin{figure}[h!]
\centering
\vspace{0.15cm}
\includegraphics[width=0.45\textwidth]{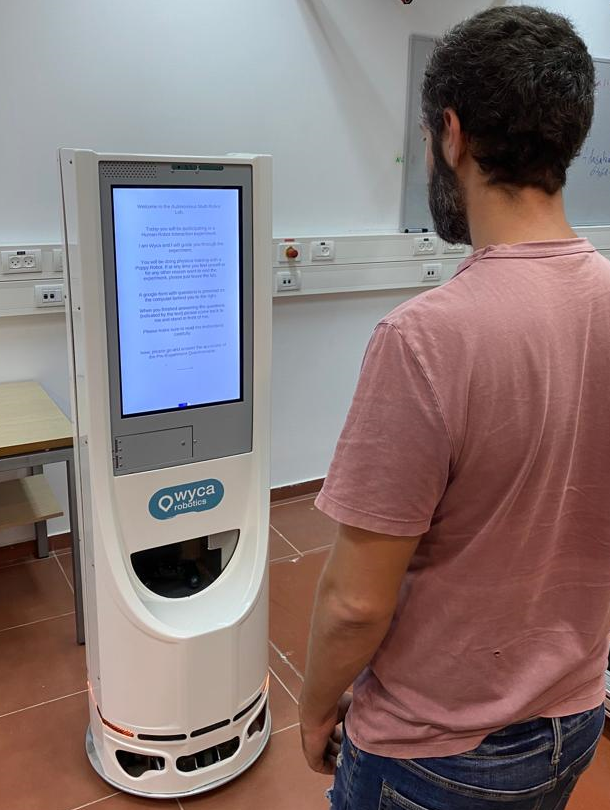}
\caption{Picture of a Participant interacting with the Wyca {\em Keylo} robot (RE).}
\label{fig:3}
\end{figure}

The RE supported us in all our tasks. 
To create our RE we used a Wyca {\em Keylo} tele-presence robot%
\footnote{\url{https://www.wyca-robotics.fr/}} (see Fig. \ref{fig:3}). 
The approximate height of the robot is 1.64 m with a circular footprint of 52 cm diameter. 
It has 24\textit{''} multi-point high touchscreen.
It has two front cameras and one rear 3D RGB-D camera. 
Note, that the robot is capable of navigating autonomously, which we did not make use of.
In future studies, the capability to move around within the experiment area is one point that distinguishes RE based experiment automation from automation of experiments via a simple computer (e.g. tablet or labtop). 
A computer screen as experimenter (Computer Experimenter, CE), would have technically sufficed for the experiment tasks we conducted via RE (e.g., automation of explanations via video and sound). However, prior work has shown that (beyond their movement capabilities) robots bring unique benefits to the table compared with CE (please see discussion section).

We scripted the full interaction in Python (i.e., the full experiment procedure) using the opencv computer vision library%
\footnote{\url{https://opencv.org/}}, 
google text to speech library (gTTS)%
\footnote{\url{https://gtts.readthedocs.io/en/latest/}}
as well as the tkinter library to display text and images on the screen%
\footnote{\url{https://docs.python.org/3/library/tkinter.html}}.
We produced a fully autonomous robot experimenter, that required us only to turn it on once in the begging of a trial session (it switches automatically from participant to participant).
Note, that we opted for a scripted interaction rather than e.g., a large language model (LLMs) to standardize the interaction (and not have any randomness in the text). 
Even to the same prompt, LLMs might produce different answers -- which is exactly what we want to reduce in our study.
Note also, that a robot experimenter has the benefit of introducing varying experimental conditions in a controlled manner -- which we did not make use of in our study.

During a trial of our experiment, the robot, placed centrally facing the door would wait until it identified a human face, which (if seen for enough consecutive frames) would trigger a displayed and spoken text.
This would happen four times, in the following order:
When the participants entered the lab it would trigger the first text -- 1. Introduction and Pre-Experiment Questionnaire.
When participants finished the questionnaire and came back to the robot it would trigger the second text -- 2. Training experiment explanation.
After participants completed the training experiment, they would come back to the robot and the third text was triggered -- 3. Internal post experiment questionnaire.
Finally, upon seeing the participants a last time after completion of the Questionnaires, it would trigger the last text -- 4. Farewell message.
When the second text is triggered it also starts the training experiment (starts the {\em Gymmy} robot). 
When the fourth text is triggered it starts a timer, which at its end resets the RE.
When a text is finished, the RE waits for a certain time before the next text can be triggered. 
The RE instructed the participants to return to it, after the questionnaires and after the training experiment -- in order to trigger the next phase.

\subsection{Participants}
A total of 33 undergraduate industrial engineering students participated in the experiment.
For their participation, which was voluntary, they were compensated with one bonus point to their grade in a course.
The experiment was publicized via the course page and open for registration to all.
Participants were all aged 24-30 years old (mean: $26.7 \pm 1.29$), 14 male and 19 female. 
The participants were randomly split into two groups (HE, RE). 
Group HE consisted of 18 students (6 male, 12 female and age: $26.61 \pm 1.42$) and group RE consisted of 15 students (8 male, 7 female and age: $26.8 \pm 1.15$). 
Prior to the experiment they only received the information that they would participate in a physical training HRI experiment (with the {\em Gymmy} robot).
The study design was approved by the Departmental Human Subjects Research Committee.
Most students had no or very limited experience with robots and (HRI) experiments.

\subsection{Data Collection and Analysis}
Subjective measures were collected via the questionnaires.
Objective measures were collected by quantitative analysis of video and notes manually taken during the experiment.
Preliminary background data of the participants was collected using the technology adoption propensity index (TAP) \cite{ratchford2012development} and negative attitude towards robots scale (NARS) \cite{nomura2006measurement}.
The TAP questionnaire was divided in two parts.
The first part contained questions about the frequency of use of technology.
The second part contained questions about whether new technologies would give the participants more control over their daily lives as well about the ease of learning to use new technologies.
Analysis was performed according to the recommendations of \cite{ratchford2012development}.
For NARS we took the sum of several questions within the questionnaire, according to the groups defined in \cite{nomura2006measurement} and then evaluated the average of each group (as suggested in \cite{nomura2006measurement}).
Further, we collected data with a post experiment questionnaire (internal), which we adopted from \cite{krakovski2021gymmy}.
Participants were asked to rate the HRI experiment (in Likert scale between one and five).
The subjective measures (internal post experiment questionnaire) were based on the Technology Acceptance Model \cite{davis1989perceived}:
{\em perceived usefulness, ease of use, attitude towards the robot and intention to use the robot}.

Questionnaires included a manipulation check, by introducing questions asking to answer with a certain value. 
This was conducted three times: 
in TAP, in NARS and in the internal post experiment questionnaire.

A last post experiment questionnaire aimed to evaluate the overall experiment (external). 
Participants were asked to rate the whole experience inside the lab -- including the interactions with the experimenter and the explanations received.
Its subjective measures were:
{\em cognitive load, comfortability, enjoyment, fluency of interaction, safety, trust and understandability}.

Finally, the objective measures, which were collected via video and notes were:
{\em success rate, time-duration of the experiment, number of times participants asked the experimenter questions, number of errors (textual, procedural)}. 
And, {\em if participants were waiting for explanations to finish or not} -- i.e., if they started moving on while explanations were given.
Under errors we collected:
textual errors -- i.e, if experimenter made a mistake giving the experiment explanation or deviated from the predefined text (e.g., swapped the order of two points);
procedural errors -- i.e., if HE made a mistake or RE ran into a bug during the experiment (e.g., RE triggering an experiment phase wrongly or HE missing a step, like the filling of a questionnaire).
During the experiment, the unnoticed observer noted all above occurrences watching the experiment through the video feed.
Finally, we went over the video material and notes together to evaluate the quantitative measures.
Note, that for a single experimenter, introducing a RE, can free up time to make and note important observations during the experiment.

The data collected was first checked for normality using a Kolmogorov–Smirnov (KS) test.
One-way analysis of variance (ANOVA) was conducted on the measures which were normal.
For non normal data, the Mann-U-Whitney test for between the groups design was conducted.
The descriptive statistics, mean and standard deviation were computed for normal data and median was calculated for not normal data.
For all evaluations, there was more than one question for each dependent variable (subjective measures) and the average of the questions was computed for each participant (resulting in a score between one and five).

The important independent variables were only the groups (i.e., group HE and RE). 
We are not emphasising on gender because the literature is inconclusive towards its effect \cite{flandorfer2012population,saunderson2020investigating,krakovski2021gymmy,Shik2022} and it is not what we seek to measure in our study. 

\section{RESULTS}
\subsection{Pre Experiment Questionnaire}
The data collected in TAP on frequency of usage was normally distributed (D = 0.128,p = 0.6445). 
Further, results revealed no statistically significant difference between groups (group HE vs RE – F(1) = 0.637, p = 0.43).
The TAP control and learning parts were also normal (D = 0.211,p = 0.104). 
There was no statistically significant difference between groups (group HE vs RE F(1) = 0.043, p=0.838).
The manipulation check employed within the questionnaires yielded that 72.72\% of all (HE and RE) respondents gave correct response. 
61.1\% and 86.67\% of the responses were correct for group HE and RE respectively.

The NARS questionnaire was found to be normally distributed (D = 0.144,p = 0.494). 
The negative perception about robots in both groups (F(1) = 0.045,p = 0.834) was without statistically significant differences.

The employed manipulation check found that 90.91\% of participants responded correctly. 
In group HE 88.89\% and in group RE 93.33\% of the responses were correct. 
Note, that the success rate between the second and first manipulation check in the pre-experiment questionnaire improved and that it is generally better when the RE is present.

\subsection{Internal Post Experiment Questionnaire}
There was no statistical significant difference between both groups (see Table \ref{tab:2}).
Nevertheless, we give descriptive statistics in Table \ref{tab:1}.
Having a human or a robot experimenter did not result in any statistically significant differences in the subjective result of the training experiment, according to our measures.
The training experiment itself is generally perceived as easy and the attitude is positive. 
The intention of use is low as the training was a very simple upper body exercise and as such too easy for young adults (as noted above).

\begin{table}[h!]
    \centering
    \resizebox{\columnwidth}{!}{%
    \begin{tabular}{|c|c|c|c|c|}
    \hline
          & Perceived usefulness & Ease of use& Attitude towards robot & Intention to use  \\ \hline
         Group HE & $3.06 \pm 1.30$ &$4.16 \pm 0.38$ &$3.67 \pm 0.61$ &$3.17 \pm 1.38$ \\
         Group RE & $2.46 \pm 1.30$ &$4.38 \pm 0.40$ & $3.33 \pm 0.54$&$2.73 \pm 1.33$ \\
    \hline
    \end{tabular}
    }
    \caption{Descriptive statistics for training experiment.}
    \label{tab:1}
\end{table}

\begin{table}[h!]
    \centering
    \resizebox{\columnwidth}{!}{%
    \begin{tabular}{|c|c|c|}
    \hline
    Subjective metric & KS Test &Group (HE vs RE) \\ \hline
    Perceived usefulness & D = 0.1849, p=0.2091& F(1) = 1.651, p= 0.209\\
    Ease of use & D = 0.1818, p = 0.2251 & F(1) = 2.639, p= 0.115\\
    Attitude towards robot & D = 0.1481, p = 0.4635 & F(1) = 2.76, p= 0.107\\
    Intention to use & D = 0.2306, p = 0.0597 & F(1) = 0.809, p= 0.376\\
    \hline
    \end{tabular}
    }
    \caption{Statistical test conducted on the subjective measure of the training experiment.}
    \label{tab:2}
\end{table}

\subsection{External Post Experiment Questionnaire}
\begin{figure}
    \centering
    \includegraphics[width=0.5\textwidth]{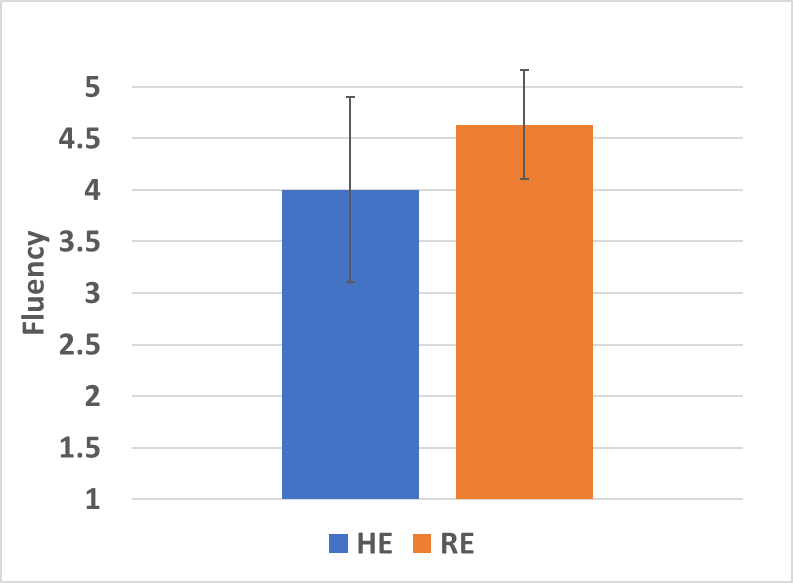}
    \caption{Plot of the mean and standard deviation for fluency.}
    \label{fig:4}
\end{figure}

The mean and standard deviation in case of cognitive load, understandability and fluency (all normally distributed) for each group are reported in Table \ref{tab:3}. 
In the cases of comfortability, enjoyment, safety and trust (not normally distributed) the median for each group is reported in Table \ref{tab:3}.
We found one statistically significant difference, which was in fluency (see Table \ref{tab:3}). 
Participants of group RE found that the experiment was more fluent then the participants in group HE.
See Fig. \ref{fig:4} for a comparative plot of the descriptive statistics for fluency.

\begin{table}[h!]
    \centering
    \vspace{0.2cm}
    \resizebox{\columnwidth}{!}{%
    \begin{tabular}{|c|c|c|}
    \hline
     & Group HE & Group RE\\
     \hline
      Cognitive Load ($X \pm S$ ) & $1.78 \pm 0.81$ & $1.8 \pm 0.77$\\
      Comfortability (median) & 4.5& 4.5\\
      Enjoyment (median) & 4 & 4\\
      Safety (median) & 4.75 & 4.5\\
      Trust (median) & 5 & 5\\
      Understandability  ($X \pm S$) & $4.26 \pm 0.51$ & $4.24 \pm 0.53$\\
      \textbf{Fluency} ($X \pm S$) & $\bm{4 \pm 0.9}$ & $\bm{4.63 \pm 0.53}$ \\
      \hline
    \end{tabular}
    }
    \caption{Descriptive statistics of subjective measures in the external post experiment questionnaire.}
    \label{tab:3}
\end{table}

\begin{table}[h!]
    \centering
    \resizebox{\columnwidth}{!}{%
    \begin{tabular}{|c|c|c|}
    \hline
     Subjective metric & KS Test &Group (HE vs RE)\\ \hline
       Cognitive Load & D = 0.20, p = 0.13 & F(1) = 0.07, p = 0.794\\ 
        Comfortability & D = 0.24, p = 0.04 & W = 105.5, p = 0.27\\
        Enjoyment & D = 0.25, p = 0.03 & W =144, p = 0.74\\
        Fluency & D = 0.22, p = 0.06 & \textbf{F (1) = 5.26, p = 0.03}\\
        Safety & D = 0.28, p = 0.008 &  W = 130.5, p = 0.87\\
        Trust &  D = 0.41, p = 0.001 &  W = 137, p = 0.95\\
        Understandability & D = 0.23, p = 0.06 & F(1) = 0.007, p = 0.94\\
        \hline
    \end{tabular}
    }
    \caption{Statistical test conducted on the subjective measures of the external post experiment questionnaire}
    \label{tab:4}
\end{table}

\subsection{Quantitative Results}
All results (descriptive statistics) are reported in Table \ref{tab:5}.
The data for time duration was normally distributed (D= 0.1201, p=0.7274), with no statistically significant difference between the groups (F(1) = 1.278, p = 0.267).
The longest trial in HE was 1020 seconds and the shortest 600 seconds long. 
In the RE group the shortest trial was 610 seconds while the longest was 920 seconds long.
In group HE questions were asked (30 in total) and the human experimenter frequently made textual errors (approx every other trial).
HE made three while RE only one procedural error.
The HE errors were mostly forgetting the internal questionnaire before the training. 
Twice the HE remembered before starting the Gymmy, once the observer had to intervene to safe the trial.
In the RE error case, the participant stayed in the field of view of the experimenter robot, which triggered a text at the wrong time.
Again, observer intervention prevented possible failure of the experiment.
I.e., we had to reset the current phase.
Lastly, we observed that participants would be more patient with the robot experimenter than with the human experimenter, i.e., they would wait for explanations to finish.

\begin{table}[h!]
    \centering
    \resizebox{\columnwidth}{!}{%
    \begin{tabular}{|c|c|c|}
    \hline
     & Group HE & Group RE\\
     \hline
      Success Rate (\%) & 100 & 93.3 \\
      Time ($X \pm S [s]$) & $782 \pm 115$ & $742 \pm 79$\\
      Questions ($X \pm S [s]$) & $1.667 \pm 1.33$  & 0\\
      Textual Error (\#) & 8 & 0\\
      Procedural Error (\#) & 3  & 1 \\
      Patience (\%) & 38.9  & 66.7 \\
      \hline
    \end{tabular}
    }
    \caption{Descriptive statistics of objective measures.}
    \label{tab:5}
\end{table}

\section{DISCUSSION and CONCLUSION}
In the first subsection we discuss the results of our experiment -- the comparison of the human led with the robot led experiment.
What are the effects of letting robots run an HRI experiment? 
Are there clear benefits and drawbacks? 

The second subsection discusses a major benefit of robot led experiments -- the replicability and validity of experiments.
We follow the result-based discussion by giving our own observations as a further qualitative measure and discuss the work's limitations.

\subsection{Robots in addition to Human Experimenters}
The experiments were successful, with only four procedural error cases in 33 experiments (three were with the human and one with the robot experimenter; none of the error cases fully invalidated an experiment). 
This shows, it is clearly possible to utilise robots to conduct the repetitive tasks in an HRI experiment.
Our results show no obvious downsides aside from robot errors however unveiled several benefits.

The participants felt as safe and professionally guided with a robot experimenter as they did with the human experimenter. 
The difference between both groups for cognitive load, understandability and trust was not statistically significant as well.
In subjective fluency, the robot experimenter outperforms the human experimenter.
Participants would ask the human experimenter many questions.
No participant attempted to direct questions to the robot experimenter.
Note that participants were not informed of the level of autonomy of the robot.
The nature of the questions and the success rates clearly indicate that there was no need to ask and in fact, asking may be counter-productive within our specific experiment.
This opens the more general question of when it is beneficial and when not for participants to ask questions. 
While a case can be made that it is better that participants ask until they understand (complex) experiment explanations, results for our simple experiment showed a different picture.
As our task was simple, questions seemed at best unnecessary. 
This is reflected when one compares the performance of RE vs HE groups on the manipulation check.
Many participants asked the HE on how to answer the manipulation check.
Asking questions did not result in better performance -- it rather seemed to slow down the experiment and made participants uncertain.
Interestingly, participants would allow the robot experimenter to finish explaining in twice as many cases as they would allow the human experimenter to do so.
Finally, we want to note that experiments led by robots might fail due to errors in robot hard and software.
In such cases it is good to have an observing and possibly intervening human experimenter present. 
Of course, error and intervention would influence the results of the experiment (e.g., by introducing human-human influence) -- and we think it should be defined beforehand when a trial has to be invalidated.
Intervention for the RE occurred only once during all our experiment.
As we were able to intervene and restart the current phase of the experiment within just a few seconds, we did not invalidate the trial.
In comparison to that, we had to instruct the human experimenter several times to stick to the presented structure, text and procedure between trials.
With HE, most mistakes were made at the beginning of running experiments (e.g., novelty and excitement) and at the very end (e.g., focus and exhaustion).
The fact that participants would let the robot finish its explanations but not the human experimenter astonished us.

\subsection{Replicability and Validity}
One of the major benefits of conducting an experiment with a robot experimenter, is that they can be replicated at later times and at different locations with relative ease and in a very precise way. 
This is, under the constrained that the same or similar robotic platforms and technologies are present.
If experiments are largely led by robots, they can be easily shared between research groups and within the community and exactly replicated (to the point of the exact same robot model, voice, behaviour and motion). 
The scripts make sure that the experiment protocol is followed. 
Unwanted human variability is reduced (actually removed from any robot performed task).
Interaction between two humans (participant and experimenter) is two sided biased, while the interaction between participant and robot is one-sided biased (the robot acts towards all participants in the same way and does not care how it is perceived).
We believe that this should increase the general validity of the experiments.

\subsection{Limitations and Future Work}
The recruited participants were engineering students that have high exposure to technology.
Previous research have suggested that, the perception about robots changes with cultural differences \cite{korn2021understanding} and age of the participants \cite{kumar2022politeness}. 
In \cite{akalin2021robot}, the authors argued that different educational background of the participants influences the interaction with robots.
Therefore, it is possible, that replication of the same or similar studies will yield different outcomes when the background of the participant changes.
A further limiting factor is posed by the robot experimenter itself.
The type of robot (e.g., size and embodiment) may have an influence on the perception and thus on the experiment itself.
Studies which require proximate working with robots (like human-robot collaboration with a manipulator as in \cite{kshirsagar2020robot}), would have safety concerns and thus would need robot skills to take care of this (e.g., shut off when danger of collision is detected).
Also, the whole interaction was scripted by us and did not allow for any reactive behavior. 
Participants did not ask the robot questions, and could not have done so but future versions could include a reactive component that allows for more elaborate interaction with the robot(s).
Furthermore, robotic systems are bound to make mistakes \cite{honig2018understanding}. 
Future robotic experimenters need to include methods for error resolution that function without human intervention, to increase the robustness of the process.
Additionally, one could argue that we could have used a computer with a screen and camera (computer experimenter CE) instead of the experimenter robot. 
Indeed, the technical functionality of leading this specific experiment was independent of the fact that it is a robot.
However, one can easily come up with cases where an experimenter (and thus also an experimenter robot) would need to move around with the participant -- preventing the use of a computer to replace the experimenter.
Moreover, previous research showed that robots are better accepted than virtual agents and that the physical presence has a positive effect on users \cite{li2015benefit,kumar2022politeness, fasola2013socially}.


\end{document}